\documentclass[sn-mathphys,Numbered]{sn-jnl}

\usepackage{graphicx}%
\usepackage{multirow}%
\usepackage{amsmath,amssymb,amsfonts}%
\usepackage{amsthm}%
\usepackage{mathrsfs}%
\usepackage[title]{appendix}%
\usepackage{xcolor}%
\usepackage{textcomp}%
\usepackage{manyfoot}%
\usepackage{booktabs}%
\usepackage{algorithm}%
\usepackage{algorithmicx}%
\usepackage[noend]{algpseudocode}%
\usepackage{listings}%

\usepackage{algorithm}

\newcommand{\PP}{\mathbb{P}}
\newcommand{\Dd}{\mathcal{D}}
\newcommand{\Pp}{\mathcal{P}}
\newcommand{\Jj}{\mathcal{J}}
\newcommand{\Ss}{\mathcal{S}}
\newcommand{\ww}{\mathrm{w}}
\newcommand{\xx}{\mathrm{w}}
\newcommand{\wm}{\mathfrak{w}}

\newcommand{\Pf}{\mathbf{P}}
\newcommand{\Qf}{\mathbf{Q}}
\newcommand{\Kf}{\mathbf{K}}
\newcommand{\Vf}{\mathbf{V}}
\newcommand{\Wf}{\mathbf{W}}
\newcommand{\Xf}{\mathbf{X}}
\newcommand{\Yf}{\mathbf{Y}}
\newcommand{\Xft}{\widetilde{\mathbf{X}}}
\newcommand{\ind}{1\hspace{-2.1mm}{1}}
\DeclareMathOperator*{\argmax}{arg\,max}

\usepackage[textsize=tiny]{todonotes}

\theoremstyle{thmstyleone}%
\theoremstyle{thmstyletwo}%
\newtheorem{example}{Example}%

\theoremstyle{thmstylethree}%

\begin{document}

\title[Natural Language Processing for Financial Regulation]{Natural Language Processing for Financial Regulation}


\author*[1]{\fnm{Ixandra} \sur{Achitouv}}\email{ixandra.achitouv@cnrs.fr}

\author[2]{\fnm{Dragos} \sur{Gorduza}}
\equalcont{These authors contributed equally to this work.}

\author[3]{\fnm{Antoine} \sur{Jacquier}}
\equalcont{These authors contributed equally to this work.}

\affil*[1]{\orgdiv{Department of Mathematics}, \orgname{Imperial College London and CNRS, Complex Systems Institute of Paris Île-de-France}}

\affil[2]{ \orgname{Oxford Man Institute of Quantitative Finance}}

\affil[3]{\orgdiv{Department of Mathematics}, \orgname{ Imperial College London and the Alan Turning Institute}}

\abstract{This article provides an understanding of Natural Language Processing techniques in the framework of financial regulation, 
more specifically in order to perform semantic matching search between rules and policy when no dataset is available for supervised learning. 
We outline how to outperform simple pre-trained sentences-transformer models using freely available resources and explain the mathematical concepts behind the key building blocks of Natural Language Processing.}

\keywords{Language models, Financial Regulation, Natural Language Processing}



\maketitle


\section{Introduction}

Over the past ten years, modern natural language processing models have revolutionised the field of artificial intelligence, transforming how computers understand and generate language. 
ChatGPT is one key example of how AI technology is becoming increasingly important in generating more and more precise human-like responses to a wide variety of problems, including business ones.
These advancements have significant implications for the financial industry, where vast amounts of regulatory data and policies must be analysed and understood in order to comply with laws and regulations.

Indeed, Natural Language Processing (NLP) has gained traction in financial regulation as a  solution for managing and interpreting complex regulatory information. 
The ability to effectively process and analyse large amounts of regulatory text is crucial for compliance officers, risk managers, and other financial professionals. 
However, traditional methods of automatic text processing including keyword searches and dictionaries are both inefficient to run at the scale required by the amount of files to treat, and costly to set up since they require system experts to design dictionaries on a case-by-case basis. Moreover, they are also prone to errors due to lack of coverage which can expose the user to costly regulatory violations.
Modern NLP techniques have emerged in recent years, mostly based on deep neural networks and after the seminal breakthrough~\cite{Vaswani} by Vaswani, Shazeer, Parmar, Uszkoreit, Jones, Gomez, Kaiser and Polosukhin, who incorporated the attention mechanism to learn latent semantic links between words in a sentence more accurately. 
These improvements, along with orders of magnitude increases in the availability of datasets and computational power have led to large language model (LLM), a type of model known for its ability to achieve general-purpose language understanding and generation such as OpenAI's GPT (used in ChatGPT), Meta's LLaMa or Google's PaLM. 
One may however wonder whether LLMs are actually required for simple NLP tasks such as semantic search, 
especially due to their costs, 
instead of free open resources. 

In the present work, we perform a semantic search applied to laws and regulation using a so called domain adaptation technique with limited amount of data, bearing in mind limited available resources from small and medium enterprises.
We also recall the historical NLP approaches and explain the leading concepts in this field. 
In particular, we focus on semantic search, which naturally leads to the encoding concepts of a sentence. 
We thus aim to fill the gap between well-known NLP concepts and the mathematical frameworks behind.

This paper is organised as follows: 
Section~\ref{sec1} is a short historical review on encoding methods of a sentence into a mathematical object as well as key machine learning advances. 
In Section~\ref{sec2} we  describe the task of semantic textual similarity, the different techniques at hand and illustrate them with pseudo codes. 
In Section~\ref{sec:applidata}, we apply these techniques concretely to financial regulation problems to improve semantic search matching. 
Section~\ref{sec3} draws conclusion from this analysis.

\section{Historical overview of word embedding approaches}\label{sec1}

Semantic vectors of words are high-dimensional vectors, and
pairwise distances between two vectors reflect the similarity of words.
We briefly review the main literature on this topic, clarifying the notions of `similarity', `meaning' and `correlation between words'.

\subsection{Deterministic approaches based on word counts}
The roots of major modern NLP techniques can be traced back Firth' quote~\cite{firth1957}
\emph{`You shall know a word by the company it keeps'}. 
Word co-occurrence statistic is a simple deterministic approach to obtain a semantic vector representation of words. 
Latent Semantic Analysis (\texttt{HAL})~\cite{LSA} and Hyperspace Analogue to Language (\texttt{HAL})~\cite{HAL} are historically the two methods that built embedded semantic vectors derived from the statistics of word co-occurrence. 
LSA derives its vectors from collections of segmented documents, 
while \texttt{HAL} makes use of unsegmented text corpora. 
For segmented documents, a document-term matrix is used to describe the frequency of terms that occur in a collection of documents such that rows correspond to the documents and columns correspond to the terms. Each $(i,j)$ cell, then, counts the number of times word~$j$ occurs in document~$i$.
Later, Rohde, Gonnerman and Plaut~\cite{Rohde2005AnIM} improved the \texttt{HAL} method introducing the Correlated Occurrence Analogue to Lexical Semantics (\texttt{COALS}) algorithm, which achieves considerably better performance through improvements in normalisation and other algorithmic details. 

\subsubsection{Co-occurrence \& Singular Value Decomposition}\label{subsec_co-occ}
The deterministic approach proposed in~\cite{Rohde2005AnIM} is one of the first methods to achieve good and consistent results in predicting human similarity judgements in words embedding. 
It is a two-step approach:

\emph{Step 1: produce a co-occurrence matrix.}
For each word~$w_1$, count the number of times another word~$w_2$ occurs in close proximity 
to~$w_1$. 
Here 'close proximity' (or 'windows size') is a hyperparameter indicating the number of words to consider before/after the central word~$w_1$. 
\texttt{HAL} suggested to use a 4-word window while \texttt{COALS} proposes 10-word windows. 
The counting is done using a weighting scheme whereby if~$w_2$ is adjacent to~$w_1$ it receives the maximal weight,
and the weight is proportional to the length separation of the central word~$w_1$ (ramped window). 
This results in an $N\times N$ co-occurrence matrix where~$N$ represents the number of words in the whole corpus. 
One could in principle stop here, where each row of the co-occurrence matrix can thus be associated with a word. 
With this procedure, \texttt{HAL}~\cite{HAL} demonstrated semantic similarity between any desired pair of words. A few tricks were also proposed in the computation of the co-occurrence matrix, for instance putting a threshold to the number of co-occurrence count to remove frequent words such as `the' or `a'. 

\emph{Step 2: Dimension reduction of word vectors.}
From the co-occurrence count of words the resulting embedded vectors should have at least the size of the words in the corpus\footnote{In fact it has a~$2N$ dimension in \texttt{HAL} where for every word in the target vocabulary, there is both a row and a column containing relevant values. 
For instance, the row may contain co-occurrence information for words appearing before the word under consideration, while the column contains co-occurrence information for words following it. 
This (row,column) pair may be concatenated so that, given an $N\times N$ co-occurrence matrix, a co-occurrence vector of length~$2N$ is available. 
In \texttt{COALS} the distinction between left/right of the central word is ignored.}. 
In \texttt{HAL}, the authors proposed to keep a relatively small number of principal components of the co-occurrence matrix. 
For instance reducing the vector size by eliminating all but the~$k$ columns with the highest variance. In~\cite{Rohde2005AnIM} the reduction of dimension is based on Singular Value Decomposition (SVD) of the normalised co-occurrence matrix (this was also proposed in for latent semantic analysis in~\cite{LSA}).

\subsection{Machine Learning}
Over the last decade, Machine Learning algorithms contribution to NLP have brought significant progress. 
Among these, \texttt{Word2vec} in 2013 by Mikolov, Chen, Corrado and Dean~\cite{Mikolov2013}, 
\texttt{Glove} in 2014 by Pennington, Socher and Manning~\cite{GLOVE}, \texttt{BERT} in 2018 by Devlin, Chang, Lee and Toutanova~\cite{BERT}
and \texttt{SBERT} in 2019
by Reimers and Gurevych~\cite{SBERT}
are the most popular ones,
which we shall use in this project. 
The key idea is to perform a non-linear regression where the coefficients are adjusted by minimising a loss function (or maximising a likelihood function). 
In the case of NLP, the coefficients/weights are the embedded vectors. 

\subsubsection{\texttt{Skip-gram} and \texttt{CBOW}}
To understand the \texttt{Skip-gram} model, we briefly review some fundamental concepts of probabilistic language modelling where the goal is to compute the probability $\PP(\ww)$
of a sequence of words $\ww:= (w_0,\ldots, w_n)$ occurring;
we also write $\ww_{i:j}:= (w_i,\ldots, w_j)$ and omit~$i$ whenever $i=0$.
A similar goal is to compute the conditional probability of an upcoming word $\PP(w_j|\ww_{j-1})$. 
The computation of these probabilities rely on conditional probabilities where the general chain of rule can be expressed as 
$$
\PP(\ww) = \PP(w_0) \PP(w_1|w_0) \PP(w_2|\ww_{:1})\cdots\PP(w_n|\ww_{:n-1}).
$$
To estimate this probability we count from a corpus
$$
\PP(w_j|\ww_{:j-1})=\frac{\# \ww_{:j}}{\#\ww_{:j-1}}.
$$
For sentences with more than a few words, the probability that a complex sentence repeats itself often enough to be able to compute statistical properties is unlikely. To overcome this, 
one usually assumes a pseudo-Markovian approximation 
(the \texttt{$k$-gram} model) \footnote{strictly speaking it is not Markovian for $k\neq 1$ because we keep a memory of the~$k$ previous words}, whereby $\PP(w_j|\ww_{:j-1}) \approx \PP(w_j|\ww_{j-k:j-1})$ for some $k\le j$. 
For instance for $k=1$ we get the properly Markov \texttt{bi-gram} model $\PP(w_j|\ww_{:j-1}) \approx \PP(w_j|w_{j-1})$. 
Given a sentence, computing for each word the probability that it occurs given the previous words is similar to filling a normalised co-occurrence matrix. 

\texttt{Skip-gram} models are generalisations of \texttt{$n$-gram} models, in which the components do not need to be consecutive, 
but may leave gaps that are skipped over~\cite{skipgram,Guthrie_acloser}. 
For instance, the adjacent words for the sentence: \emph{`I hit the blue ball'} are $\{\emph{I hit}, \emph{hit the}, \emph{the blue}, \emph{blue ball}\}$ for a
\texttt{bi-gram} and 
$\{\emph{I hit}, \emph{I the}, \emph{I blue}, \emph{hit the}, \emph{hit blue}, \emph{hit ball}, \emph{the blue}, \emph{the ball}\}$
for a \texttt{2-skip-bi-gram}.
Finally, in \texttt{Word2vec}~\cite{Mikolov2013a},
the definition of the \texttt{Skip-gram} architecture is to predict surrounding words given a specific word, which formally corresponds to computing $\PP(\ww_{n-j:n+j}|w_n)$. while the Continuous Bag of Words (\texttt{CBOW}) architecture predicts the current word based on the context: $\PP(w_j|\ww_{:j-1})$.  These two models are illustrated in~\cite[Figure~1]{Mikolov2013a}.

\subsubsection{Link with the embedding vectors}\label{secemvec}
In the \texttt{Skip-gram}  originally presented in~\cite{Mikolov2013a}, 
the training objective is to learn word vector representations that are good at predicting the \emph{close-by} words $w(t-2), w(t-1),w(t+1),w(t+2)$ from an input word $w(t)$. 
This is achieved by maximising the average log likelihood probability, equivalently by minimising the loss function
\begin{equation}\label{eqloss1}
\Jj(\Theta):= -\frac{1}{T} \sum^{T}_{t=1} \sum_{-c\le j\le c, j\ne 0} \log \PP(w_{t+j}|w_t),
\end{equation}
where~$c$ is the size of the training context, $(w_1, \cdots, w_T)$ is a sequence of training words 
and~$\Theta$ is the vector representation of words that the algorithm tries to optimise.
Note that every word~$w$ appears both as a central word (input ~$w_I$) and as a context word (output ~$w_O$) so that 
$\Theta \in \mathbb{R}^{2\wm}$ where~$\wm$ is the size of the vocabulary\footnote{the final embedded word~$w$ is simply the average of its input~$w_I$ and output~$w_O$ vector representations.}.
In the basic \texttt{Skip-gram} formulation, a softmax function is proposed to model the loss function, namely

\begin{equation}\label{eqsoft}
\PP(w_O|w_I)=\frac{\exp\left(v_{w_O} ^{\top}v_{w_{I}}\right)}{\sum_{w=1}^{\wm}\exp\left(v_{w}^{\top}v_{w_{I}}\right)},
\end{equation}
where $w_O$ and $w_I$ are the output and input words,
and, given a word~$w$, $v_{w_{I}}$ and~$v_{w_{O}}$ are its input and output vector representations. 
The loss function is then minimised via gradient descent:
the gradient of the likelihood (with respect to~$v_{w_{I}}$) can be computed easily and is equal to zero if and only if
\begin{equation}
    v_{w_{O}} = \left[\sum_{w=1}^{\wm}\exp(v_{w}^{\top} v_{w_{I}})\right]^{-1} \sum_{w=1}^{\wm} v_{w} \exp(v_{w}^{\top}v_{w_{I}}),
\end{equation}
which is the expectation of $v_{w_{O}}$ occurring given the context word $v_{w_{I}}$. 

In practice, to improve computation efficiency, a hierarchical softmax is proposed in~\cite[Equation~(3)]{Mikolov2013a}), which uses a binary tree representation such that the computing time cost reduces from~$\wm$ to~$\log(\wm)$.
An alternative technique to improve efficiency is to sub-sample~$k$ random words and compute their probabilities to appear in the context of the central words. This is known as negative sampling where $\rm \log \PP(w_{t+j}|w_t)$  in~\eqref{eqloss1} is replaced by 
\begin{equation}\label{eq:NegSampling}
\log\sigma\left(v_{w_O}^{\top}v_{w_I}\right)
   + \sum_{j=1}^{k} \mathbb{E}_{w_j\sim P_n(w)}\left[\log\sigma\left(-v_{w_j}^{\top}v_{w_I}\right) \right],
\end{equation} 
where~$\sigma(\cdot)$ is the sigmoid function and the sum runs over a random sample of~$k$ words (rather than~$\wm$) based on their frequencies around the center word. 
Thus, the task is to distinguish the target word~$w_O$ from draws from noise distribution $P_n(w)$ using logistic regression. 
The noise distribution~$P_n(w)$ in~\eqref{eq:NegSampling} is a free parameter and~\cite{Mikolov2013a} found that $P_n(w) = u(w)^{3/4}/Z$, where~$u(w)$ is a unigram and~$Z$ is just the normalisation constant, outperformed computational efficiency significantly over the uniform distributions.
This avoids summing over all words in the denominator of~\eqref{eqsoft}. In~\cite{Mikolov2013a}, typical values for~$k$ lie in the range $[5,20]$ for small training datasets, while for large datasets $k$ can be as small as $[2, 5]$.
An implementation of this model can be found at \url{https://github.com/chrisjmccormick/word2vec_commented/blob/master/word2vec.c}.  
Improvements of the \texttt{CBOW} and \texttt{Skip-gram} models have been proposed but still rely on the same fundamental idea of co-occurrence statistics of words from~\cite{GLOVE}. 

\subsubsection{Transformer and attention mechanism}

Another major step was archived using multi-layer bidirectional Transformer encoders~\cite{Vaswani} for the training architecture, and was used in \texttt{BERT}~\cite{BERT} (\emph{Bidirectional Encoder Representations from Transformers}).
A main difference with \texttt{Word2vec} is that \texttt{BERT} is a non-static model with not just one hidden layer. 
Another key difference is that there is no window-size limitation of n-words in \texttt{BERT} to consider the context of the word to encode. 
The technique employed instead is known as the \emph{attention mechanism} and is the building block of a transformer architecture.  
Transformers were originally designed for translation and they are now the state of the art architecture in NLP. 
The transformer model architecture can be seen in~\cite[Figure~1]{Vaswani}.

Transformers can be summarised as encoder-decoder tools, where a sentence is fed to the encoder and the decoder generates another sentence;
the whole system is built by stacking several layers or encoder/decoder. 
Each encoder has two sub-layers,
a multi-head self-attention layer and a position-wise fully connected feed-forward network, 
and receives a list of vectors as input, passes it into a self-attention layer, 
then into a feed-forward neural network, and finally sends out the output upwards to the next encoder.
The decoder itselg has tree layers:
in addition to the two sub-layers of the encoder layer, the decoder inserts a third sub-layer, which performs multi-head attention over the output of the encoder stack and helps the decoder focus on relevant parts of the input sentence.

The Self-attention mechanism is the building block of a Transformer and captures the relationship between the different words in a sentence.
All attention mechanisms in a Transformer are split into independent heads whose outputs are concatenated before being further processed. 
A self-attention mechanism computes a score value between a query with keys, 
and reweighs the value.
All heads' values will be used through a scale dot architecture to update the attention weight matrix.
A self-attention head (assuming one single head) can summarised as follows~\cite{phuong2022formal}:
\begin{itemize}
\item For each word $i \in \{1,\ldots, L\}$ in a sentence of length~$L$, return its positional embedding vector~$\xx_i$ of length~$d_e$, which is one of the hyperparameters. 
In~\cite{Vaswani} the base case is $d_e=512$ while for example in GPT2 it is $768$. 
The size of the embedding impacts the probability that two random embedded words are correlated or not 
(if their scalar product is close to~$\pm 1$ or to~$0$ respectively).

\item Stack each embedding into a matrix $\Xf \in \mathbb{R}^{L \times d_e }$. For instance if we consider each English word as a token, the sentence 'I hit the blue ball' has $L=5$ tokens, each of them encoded in a $512$-dimensional embedding vector. 

\item Compute the query, key and value matrices~\textbf{Q}, \textbf{K} and~\textbf{V} by projecting~$\Xf$ in the (subspace) representations of the packed queries, keys and values: 
$$
\Qf:= \Xf \Wf_{q},\qquad
\Kf:= \Xf \Wf_{k},\qquad
\Vf:= \Xf \Wf_{v}, 
$$
where  $\Wf_{q}$, $\Wf_{k}$, $\Wf_{v}$  are weights matrices of length $d_e  \times d_w$ trained during the training process (randomly initialised) and their dimensions depend on the architecture. 
For simplicity, in one single-head attention $d_w=d_e$, 
while for multiple-head ($8$ in~\cite{Vaswani}), $d_w=d_e/N_{\rm head}=512/8=64$. Multiple-head attention algorithm split multiple query, key and value `heads' in order to improve the performance of the algorithm. This split reshapes weight matrices into $\Wf_{q}^i$, $\Wf_{k}^i$, $\Wf_{v}^i$, of dimensions $d_e \times  d_w$.
In the base case of~\cite{Vaswani}, $d_w=64$ was selected after hyperparameter tuning as was the length of the embedded vectors~$d_e$. Larger values of the hyperparameters such as~$d_w$ or~$d_e$ provide better performance~\cite[Table~4]{Vaswani}, but require more computational resources. 

\item Compute a self-attention score of each word of the input sentence against the others, by taking the dot product between query and key vectors of the respective words. 
These scores are then normalised and passed into a softmax function. 
The resulting attention-weighting matrix of size $L \times L$ represents the correlation between word~$i$ and word~$j$. 
Finally, the attention function or self-attention head takes the attention-weighting matrix and multiply it by the value matrix~$\Vf$, defined as 
\begin{equation}
\rm Attention(\Qf, \Kf, \Vf )=\mathrm{softmax}\left(\frac{\Qf\Kf^{\top}}{\sqrt{d_w}}\right)  \Vf,
\end{equation}
\end{itemize}
resulting in an updated vector representation of the contextualised token~$\Vf$. 




The whole transformer itself uses~$h$ attention heads ($h=8$ in~\cite{Vaswani})
and is summarised in Algorithm~\ref{MultiheadedAttention} to allow the model to jointly represent
information from different representation subspaces at different positions. 
The reweighing “values” from the heads are passed through another densely connected layers. 
The outputs of these attention heads are concatenated into a matrix~$\mathbf{O}$ of dimensions $L \times  d_e$. This matrix is multiplied by an output weight matrix~$\Wf_{o}$ of dimension $d_e \times  d_e$. The result is ~$\Xft$, an updated representation of~$\Xf$ the input matrix of the embedded tokens with $\Xft\in \mathbb{R}^{L \times d_e }$.

\begin{algorithm}
\caption{Multiheaded Attention Layer}\label{MultiheadedAttention}
\begin{algorithmic}%
\Require $\Xf$, $\Wf_{q}$, $\Wf_{k}$, $\Wf_{v}, \Wf_{o}$%
\For{$i = 1$ to $h$} (Split $\Wf_{q}$, $\Wf_{k}$, and $\Wf_{v}$ into~$h$ heads $(\Qf_i, \Kf_i, \Vf_i)_{i=1,\ldots, h}$)
\State $\Qf_i \gets  \Xf \Wf_{q}^i$;
\qquad $\Kf_i \gets  \Xf \Wf_{k}^i$;
\qquad $\Vf_i \gets  \Xf \Wf_{v}^i$

\State $\mathrm{output}_i \gets \mathrm{Attention}(\Qf_i, \Kf_i, \Vf_i) $;
\EndFor
\State $\mathbf{O} \gets \text{Concatenate}(\mathrm{output}_1, \ldots,\mathrm{output}_h)$
\State $\Xft \gets \mathbf{O} \Wf_{o}$
\Ensure $\Xft$
\end{algorithmic}
\end{algorithm}

The transformer model has been very successful and is at the core of many machine learning applications:
for NLP, the \texttt{BERT} family~\cite{Vaswani},
in biology, the Generative Pre-trained Transformer (\texttt{GPT}) family to predict molecular structures~\cite{biotrans}, in computer vision, to extract image information~\cite{vistrans}. 
Recent models differ from their predecessors primarily by their size (\texttt{BERT} has 310 million parameters,  \texttt{GPT3} has 175 billion~\cite{NEURIPS2020_1457c0d6}, 
\texttt{PALM} has 540 billion). 
The other chief difference lies in which part of the encoder-decoder framework of the transformer architecture is leveraged. 
\texttt{BERT} uses the entire encoder-decoder while models like the \texttt{GPT} family use the only decoder.
The classical \texttt{BERT} architecture is trained on two tasks \cite{BERT}. The first is sentence entailment, where the input is one sentence and the model is asked to predict whether the question proposed as the output is logically entailed by the first or not. 
The second task is Masked Language Modelling, where a random $15\%$ subset of tokens in the entire training corpus is replaced by a [MASK] token and the \texttt{BERT} model is asked to predict the most likely candidate replacement. 
Both these pre-training approaches yield a very powerful model that can be fine-tuned on a downstream task.

\section{Semantic textual similarity search}\label{sec2}

To perform semantic similarity search, a classical approach would be to feed an input sentence or text to the \texttt{BERT} transformer network which produces contextualised word embeddings for all input tokens in the text.
Then, via a pooling layer (such as mean-pooling), the average of the contextualised word embeddings would return a fixed-sized output representation vector. 
In the \texttt{BERT}-base model the dimension is 768; similarly to the previous discussion, larger values provides better accuracy but at the cost of computational memory and time. 
This sentence/text could be directly compared with another pair using for instance cosine similarity between the two vectors.
However this would often yield poor sentence embedding compared to \texttt{GloVe}~\cite{SBERT} and to expensive computational time.
For instance in~\cite{SBERT}, 
computing the most similar sentence pairs from~$1,000$ sentences took~$65$ hours with \texttt{BERT}.
We have also tried to perform some fine-tuning and further train \texttt{BERT} on our corpus and indeed found the computational time to be significantly larger and the semantic search score high even for non matching (policy, rule) pair.  

To palliate this, \texttt{SBERT} (Sentence-BERT) uses a \texttt{BERT} architecture with siamese and triplet networks~\cite{Schroff_2015} that is able to derive semantically meaningful sentence embeddings. 
This adds a pooling operation to the output of \texttt{BERT} and fine-tune the latter by adding a training dataset with pairs of sentences.
The cosine similarity between a query vector~$\Qf$ and a policy vector~$\Pf$ can be defined as
\begin{equation}\label{eq:CosineSim}
\Ss_{c}(\Qf,\Pf):= \frac{\Qf^\top\Pf}{\Vert\Qf\Vert \Vert \Pf\Vert},
\end{equation}
which converges to one as the similarity increases. 
This allows \texttt{SBERT} networks to be fine-tuned and to recognise the similarity of sentences.  
Different versions of loss functions can be used; for example, we later use  the \texttt{all-MiniLM-L6-v2}, trained on a dataset of over one billion training pairs. 
These models can be freely accessed using Sentences Transformers (\url{www.sbert.net}) to perform text embeddings directly from a large collection of pre-trained models tuned for various tasks.

Our goal here is to link rules issued by an organisation (such as governments or financial regulators) to financial institutions policies (banks in particular).  
Financial rules are often updated and financial institutions are required to issue new policies addressing these changes. 
It is extremely challenging for large institutions to keep track of regular updates and to ensure their policies match the latest rules. 
Using NLP, we perform here semantic search between query sentences (rules) and answers (policies) written in a corpus as follows:
\begin{itemize}
\item Clean corpus and rules from meaningless symbols (bracket,  hashtag, special character, ...).
\item For each query, split the corpus into sentences or paragraphs, depending on the data and the query size.
Sentence-by-sentence comparison similarity will be more precise than pooling many sentences into paragraphs. 
\item Use a pre-trained model such as \texttt{BERT} to tokenise each word of the sentence, and map sentences to embedded vectors, usually via mean-pooling of each embedded token.  
The size of any embedded token depends on the pre-trained model.
One can also use pre-trained model such as \texttt{all-MiniLM-L6-v2} and embed the sentence directly using sentence-transformer library.
\item Use cosine similarity~\eqref{eq:CosineSim} as a metric between an embedded rule query and the embedded policies to find the best match.  By requiring that a match is positive above a threshold (say $\Ss_{c}>0.7$),
 we can easily extract from a policy corpus which sentences match the rule (if any). 
\end{itemize}
This powerful pipeline can however be improved.
Using pre-trained models to encode our sentences to vectors may sound sensible, but they are trained on generic corpus and are not familiar with domain-specific targets--such as financial regulation--with high accuracy. 
One may instead completely re-train \texttt{BERT} on a domain-specific corpus, as done in~\cite{SciBert}; 
this however requires huge amount of training data ($3.3$ billion tokens),
which is extremely costly.
Instead--and this is the state of the art-- training can be performed using an Adaptative Pre-Training, described at \url{www.sbert.net};
this was successfully applied in various domains, such as for the bio-medical language~\cite{BioBERT}. 
It can then be  archived with Sentence-Transformers frameworks and the \texttt{Hugging Face} Transformers library, specifically built for NLP applications \url{huggingface.co}.  

\subsection{Classical Domain Adaptation  (DA) Pre-Training method}\label{sec:DA}
The DA Pre-Training method can be split into two steps:
\begin{itemize}
\item Step 1: further train a pre-trained model for words embeddings such as \texttt{BERT},  \texttt{SBERT} or \texttt{all-MiniLM-L6-v2} using a context specific corpus. 
\item Step 2: fine-tune the resulting model based on an existing training dataset (such as paired sentences),  similarly to \texttt{SBERT} but on the domain-specific dataset.
\end{itemize}

Before performing Step~1, one can also add up domain-specific tokens to the existing model, 
although given the limited amount of datasets at hand, this may not always be possible.  
To perform Step~1, known as Pre-Training on Target Domain, we use the Mask Language Modelling (MLM) approach~\cite{TSDAE}, which masks a random fraction of tokens in a sentence and the transformer then tries to guess what is missing. 
Alternative approaches include Transformer-based Denoising AutoEncoder (TSDAE)~\cite{TSDAE}.
While MLM outputs a token vector (mask word), the TSDAE encoder is fed with noisy sentences which the decoder uses to predict the full original sentences. 
To perform Step 2,  we need to specify the loss function and the training set.  
Many generic training sets with labelled paired sentences are available (\url{huggingface.co/datasets}), but we do not use them since wish to target our fine-tuning on (rules, policies) pairs.  
For the loss function, we use the Multiple Negatives Ranking (MNR) Loss detailed in~\cite[Section~4]{MNR} and reviewed below.

The idea of Multiple Negatives Ranking (MNR), introduced in~\cite{MNR}, is to suggest responses to a question embedded in a vector~$\Xf$ maintaining some kind of memory throughout the course of a dialogue, which translates to finding the probability $\PP(\Yf|\Xf)$, where~$\Yf$ is the embedded vector answer.  This probability is used to rank possible answers~$\Yf$ given an input question~$\Xf$. 
Bayes' theorem and the total law of probability indicate that this probability can be expressed as 
\begin{equation}\label{Eqpdfygx}
\PP(\Yf|\Xf) = \frac{\PP(\Xf,\Yf)}{\sum_{k} \PP(\Xf,\Yf_k)}, 
\end{equation}
and the joint probability $\PP(\Xf,\Yf)$ is estimated using a neural network scoring function~$\Ss$ such that 
$$
\widehat{\PP}(\Xf,\Yf):= 
\exp\left\{ \Ss(\Xf,\Yf)\right\}.
$$
In practice the denominator in~\eqref{Eqpdfygx} (equal to~$\PP(\Xf)$) is approximated by sampling~$K$ responses from the training corpus with $\widehat{\PP}(\Xf):= \sum_{k=1}^{K} \PP(\Xf,\Yf_k)$, 
leading to the approximate probability used to train the neural network:
$$
\widehat{\PP}(\Yf|\Xf):= 
\frac{\exp\left\{ \Ss(\Xf,\Yf)\right\}}{\sum_{k=1}^{K}\exp\left\{ \Ss(\Xf,\Yf_k)\right\}}.
$$
Having a training dataset with (question, answer) pairs,  each embedded question~$\Xf_i$ is paired with the embedded answer~$\Yf_i$ and all other~$\Yf_j$ for $i\neq j$ is treated as a negative candidate for question~$\Xf_i$ (so $K-1$ negatives).  The goal of the training is to minimise the loss function 
$$
\Jj(\Xf,\Yf,\Theta):= -\frac{1}{K}\sum_{i=1}^{K} \log\widehat{\PP}(\Yf_i|\Xf_i),
$$
where~$\Theta$ gathers the word embeddings and the neural network parameters. 
To compute the score, the authors in~\cite{MNR} represent the input question tokenised into a word sequence~$\Xf$ and responses~$\Yf$ as fixed-dimensional input features, extracting $n$-gram features from each. 
During training, they learn a $d$-dimensional embedding for each $n$-gram jointly with the other neural network parameters. 
To represent sequences of words,  they combine $n$-gram embeddings by summing their values.  
This bag of $n$-grams representation is denoted as $\Psi(\Xf) \in \mathbb{R}^d$. 
Then a feedforward scoring model takes the $n$-gram representation of a question and a response, and computes a score. 
For instance in the dot-product architecture from~\cite[Figure~3]{MNR},
$\Psi(\Xf)$ and $\Psi(\Yf)$ go to  two separate $\tanh$ activation layers returning the encoded~$\Xf$ and~$\Yf$ as $\mathbf{h}_x$, $\mathbf{h}_y$ and performing $\Ss(\Xf,\Yf)=\mathbf{h}_x^{\top}\mathbf{h}_y$.  
Using the MNR Loss, the score is computed using at first the chosen pre-trained model and the default scoring is computed via cosine similarity in~\eqref{eq:CosineSim}.\footnote{see \url{github.com/UKPLab/sentence-transformers/blob/master/sentence_transformers/losses/MultipleNegativesRankingLoss.py} for details} 
The key issue and challenge for this work lies in the absence of a training dataset for policies and rules.  
While unsupervised text embedding learning performs rather poorly in learning on domain-specific concepts without fine tuning, we tested an approach to perform an Adaptative Pre-Training without using the training dataset we generated in Section~\ref{secval}. This method generates pseudo-labeling pairs for fine tuning and we describe it in the next section.

\subsection{Domain Adaptation with Generative Pseudo-Labeling}
Alternatively to the approach described above, 
we also tested the unsupervised fine-tuning with Generative Pseudo-Labeling (GPL)~\cite{GPL}.  
This method can be used to perform Step~2 from Section~\ref{sec:DA}.
In our context we wish to investigate how it improves Step~2 over a fine-tuning fitting method that uses labelled (rules, policies) pairs.
The idea behind GPL can be condensed into four steps, that we now describe.
Below, we shall use calligraphic letters to denote sets (of paragraphs for example) as opposed to standard letters for elements, for example $\Pp^- := \{P^{-}_1, P^{-}_2, \ldots\}$.

\subsubsection{GPL Step 1: Generating queries from a domain specific corpus} 
For each paragraph of the corpus, we generate queries using a $\mathrm{T5}$-encoder-decoder model~\cite{T5deco} similarly to the architecture of~\cite{Vaswani} discussed in Section~\ref{secemvec}.  
This query generator, which can easily be called again from Sentences-Transformer models, yields positive (query, answer) pairs and is summarised in Algorithm~\ref{Generator}.

\begin{algorithm}
\caption{Generator}\label{Generator}
\begin{algorithmic}
\Require input paragraph~$P^{+}$, 
vocabulary size~$V$, 
length of the output query paragraph~$L_q$
\State $G \gets \mathrm{T5}(P^{+})$
($G \in \mathbb{R}^{V \times L_q}$: matrix of probability distributions)
\State $Q \gets (\argmax(G_i))_{i = 1, \ldots L_q}$
\State $Q \gets \mathrm{Decode}(Q)$
\Ensure (Q,$P^{+}$)
\end{algorithmic}
\end{algorithm}

\subsubsection{GPL Step 2: Negative mining using dense retrieval}
This step finds sentences in the text that share many similar words. For each query or paragraph, 
there is now a positive pair.
In addition, for each of the generated queries,  GPL retrieves~$M$ negative passages that are similar but not a match to the query, which is known as a negative mining process.  
At the end of this step, we obtain triplets of positive and negative passages associated with a query. 
This step makes use of a function \texttt{DenseRetrieval} which is detailed in Algorithm~\ref{DenseRetrieval}.

\begin{algorithm}
\caption{Dense Retrieval}\label{DenseRetrieval}
\begin{algorithmic}
\Require Remaining set~$\Pp = \{P_1,\ldots, P_{|\Pp|}\}\setminus\{P^+\}$ of paragraphs; 
user query~$Q$; 
$M\leq |\Pp|$: number of negative passages;
\State $\mathrm{\Pp^{-}}, \Dd \gets \{\}, \{\}$
\State $E \gets (\mathrm{T5}(P_i))_{i = 1,\ldots,|\Pp|}$\quad (Calculate an embedding $E$ for each paragraph)
\State $E_{Q} \gets \mathrm{T5}(Q)$ \quad 
(Calculate an embedding $E_{Q}$ for from the query $Q$)
\State $\Dd \gets ({\Ss_{c}(E_i,E_Q ),E_i})_{i = 1, \ldots|E|}$
\quad (with $\Ss_{c}(\cdot)$ defined in~\eqref{eq:CosineSim})
\State $\Dd \gets \mathrm{sort}(\Dd)$ (Sort by cosine similarity)
\State $\Pp^{-} \gets (D_{1}, \ldots, D_{M})$
\Ensure $(Q,P^{+},\Pp^{-})$
\end{algorithmic}
\end{algorithm}

\subsubsection{GPL Step 3: Pseudo labelling}
Use a crossencoder \cite{humeau2019poly} to score all query-positive passage and query-negative passages.
Note that some negative passage might actually be positive pairs.
This step scores the triplets composed of 
our initial paragraph~$P^{+}$, 
our generated query based on the initial paragraph~$Q$, 
one output~$P^{-}_{i}$ from our set of negative samples~$\Pp^{-}$.

\begin{algorithm}
\caption{Pseudo Labelling}\label{PseudoLabelling}
\begin{algorithmic}
\Require $(Q, P^{+}, \Pp^{-})$, crossencoder, $M$
\For{$i = 1$ to $M$}
\State $\mathrm{margin}_i \gets \mathrm{crossencoder}(Q, P^{+}) - \mathrm{crossencoder}(Q, P^{-}_{i})$
\EndFor
\Ensure $(Q,P^{+},P^{-}_{i},\mathrm{margin}_i)_{i=1,\ldots,M}$
\end{algorithmic}
\end{algorithm}

\subsubsection{GPL Step 4: Training/Tuning the Transformer model to identify the differences between positive and negative passages,  
using a crossencoder model}
The latter compare the embeddings of the passages by generating similarity scores for both positive and negative pairs. 
Given a transformer model that we wish to train using GPL and the output tuples from the previous step, we will train the said transformer by backpropagation between the margin computed in Algorithm~\ref{PseudoLabelling} and the predicted margin we extract.
This process is optimised using an MSE loss function applied to the margin
\begin{equation}\label{eq:EqMargin}
 S_{\mathrm{MSE}}:= 
\left|X_{Q}^\top \left(X_{P^{+}} - X_{\Pp^{-}_{i}}\right)  \right|,
\end{equation}
with $ X_{Q}, X_{P^{+}}, X_{\Pp^{-}_{i}}$ the embedded vectors of the query, positive passage, and negative passage $_{i}$.

\begin{algorithm}
\caption{Training loop of Transformer using GPL}\label{Trainning}
\begin{algorithmic}
\Require $(Q,P^{+},P^{-}_{i},\mathrm{margin}_i)_{i=1,\ldots,M}$
\For{$i = 1$ to $M$}
\State $(X_{Q}, X_{P^{+}}, X_{P^{-}_{i}}) \gets \mathrm{Transformer}(Q,P^{+},P^{-}_{i})$
\State $\mathrm{predicted}_{\mathrm{sim}^{+}} \gets \Ss_{c}(X_{Q},X_{P^{+}})$ 
\quad (with $\Ss_{c}(\cdot)$ defined in~\eqref{eq:CosineSim})
\State $\mathrm{predicted}_{\mathrm{sim}^{-}_{i}} \gets \Ss_{c}(X_{Q}, X_{P^{-}_{i}})$
\State $\mathrm{predicted}_{\mathrm{margin}_{i}} \gets \mathrm{predicted}_{\mathrm{sim}^{+}}-\mathrm{predicted}_{\mathrm{sim}^{-}_{i}} $
\State $\mathrm{Loss}_i \gets S_{\mathrm{MSE}}(\mathrm{margin}_i,\mathrm{predicted}_{\mathrm{margin}_{i}})$ defined in~\eqref{eq:EqMargin})
\State $\mathrm{Output}_i \gets \mathrm{backpropagate}(\mathrm{Loss}_i)$
\EndFor
\Ensure $\mathrm{Output}$
\end{algorithmic}
\end{algorithm}

\subsection{To summarise}

Both supervised and unsupervised domain adaptation methods use a pre-trained model and further train it by comparing (query, answer) pairs or triplets (negative answer). 
These pairs and triplets are in the classical supervised method given by the dataset used to perform the domain adaptation (corpus of a specific domain), but it may also be generated.
That is the GPL approach for a fully unsupervised domain adaptation. 
Below, we explore the results of performing an Adaptative Pre-Training method and a purely unsupervised approach, using GPL. To compare our results and to perform supervised fine-tuning we need a pseudo training/validation dataset. We describe in the next section how we create it.

\section{Application to semantic matching for financial regulation}\label{sec:applidata}
\subsection{Creation of a pseudo training and validation dataset}\label{secval}

To create our validation dataset we use a mean assemble average of N pre-trained models (in what follows N=10)\footnote{
\texttt{multi-qa-mpnet-base-cos-v1},
\texttt{sentence-t5-xl},
\texttt{multi-qa-distilbert-cos-v1},
\texttt{msmarco-bert-base-dot-v5},
\texttt{all-distilroberta-v1},
\texttt{all-MiniLM-L12-v2},
\texttt{distiluse-base-multilingual-cased-v2}, 
\texttt{all-mpnet-base-v2},
\texttt{stsb-distilbert-base},
\texttt{bert-base-nli-mean-tokens}. These models can be found at \url{www.sbert.net/docs/pretrained_models.html}} for the task of semantic search between rules and policies.
We run these N sentence-transforming models on a catalogue of rules and financial policies provided by FinregE (\url{https://finreg-e.com}), a company providing clients in financial services
with a software focusing on current and future regulations, using ML and AI tools to identify and interpret regulatory requirements and to integrate compliance workflows for action and compliance management. 

The dataset of rules is publicly available from Financial Conduct Authority (FCA) Rulebook 2022. After cleaning and splitting the rules into sentences of length 200, we obtain about 50,000 sentences, composed of 21,914 rule IDs. 
The policies' dataset was obtained from  FinregE and is composed of 2,374 policies that may not cover all of the newly edited FCAR rules
Our approach is the following: 
\begin{itemize}
\item For each model, we keep the rule/policy match sentences if the cosine similarity~\eqref{eq:CosineSim} is above~$0.7$. This is an empirical choice that provides fair matches. 

\item We keep pairs that have been identified by a number of models greater than~$\sqrt{N}$.
Given that the standard deviation of shot noise is equal to the square root of the average number of events N, hence if models are uncorrelated, selecting a signal to noise match above one sigma should correspond to a~$\sqrt{N}$ cut. 

\end{itemize}

\begin{figure}
\begin{center}
\includegraphics[scale=0.4]{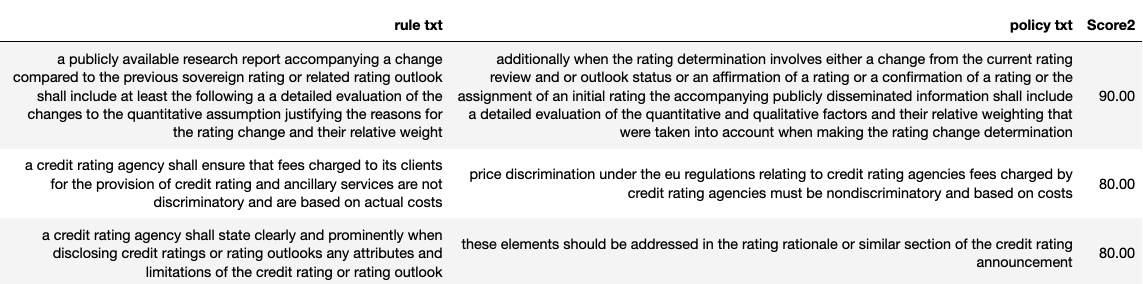}
\caption{Sample of the semantic text search output between rules and policies. 
The score corresponds to the percentage of models that have identified the same match.}\label{Fig2}
\end{center}
\end{figure}

We end up with~$1,760$ matching (rule, policy) pairs out of which we keep~$1,408$ pairs to fine-tune our model (Section~\ref{secFT}) and~$352$ for the validation dataset.  
In Figure~\ref{Fig2} we can see a sample of our validation dataset results.
This dataset will be good to identify pairs that share high similarities but will not be able to pick up on subtle pairs. This is why it is not an unbiased dataset for fine-tuning purposes but we proceed anyway due to the absence of hand labelled pairing between rules and policies.

\subsection{Domain Adaptation Pre-Training method}\label{secFT}
The corpus we use for further pre-training 
(Section~\ref{sec:DA}-Step~1) 
are rules from the FCA Rulebook. We use a full word mask and we use the \texttt{DataCollatorForLanguageModeling} which is a class of transformers to perform the MLM training\footnote{see \url{huggingface.co/docs/transformers/main_classes/data_collator} and \url{www.sbert.net/examples/unsupervised_learning/MLM/README.html}}.  
Then we run Step~2 of Section~\ref{sec:DA} with the MNR loss using our training dataset matching (rule, policy) pairs.
We perform Step~1 and Step~2 using both \texttt{BERT} and \texttt{all-MiniLM-L6-v2} pre-train model to further train in our domain-specific corpus.  

\subsubsection{Results}
In the absence of an unbiased validation dataset, it is difficult to perform proper model comparison. 
We do so by rating two scores. 

\begin{itemize}
\item Score 1: margin between matching and random pairs. For all the rules in our validation datasets we use the model considered to compute the cosine similarity $\Ss_c(R_i, P_i)$ between rule~$i$ and policy~$i$ (the matching pair) and the cosine similarity $\Ss_c(R_i, P_j)$ between rule~$i$ and a random policy~$j$, with $j\neq i$. 
Then we compute 
\begin{equation}
\Ss_1:=  \frac{1}{N} \sum_{i=1}^{N}
\Big\{\Ss_c(R_i, P_i) - \Ss_c(R_i, P_j)\Big\},
\end{equation}
where~$N$ is the length of the validation dataset (here $N=352$).
\item Score 2: For each rule~$R_i$, 
we compute the highest similarity scoring policy 
$P_i^* := \argmax_{j=1,\ldots,N}\Ss_c(R_i, P_j)$.
If $P_i^* = P_i$, we then increment~$\Ss_2$ by one and then divide by~$N$ to get the fraction of `correct' matches per model,
namely
$$
\Ss_2 := \frac{1}{N}\sum_{i=1}^{N}\ind_{\{P_i = \argmax_{j=1,\ldots,N}\Ss_c(R_i, P_j)\}}.
$$
\end{itemize}

For a perfectly tuned model and for an unambiguous validation dataset, $S_1$ and $\Ss_2$ should be close to~$1$ since $\Ss_c(R_i, P_i)$ approaches~$1$ and $\Ss_c(R_i, P_j)$ approaches zero. 
In practice, since we split rules and policies in sentences, it may happen that the same sentence occurs in different policies. Therefore a match between~$R_i$ and~$P_j$ (with $j\ne i$) is not necessarily a wrong match.
In addition in each rule, some sentences are quite generic and are addressed by many policies.
Our score does not account for these and is therefore just a benchmark to test the impact of the domain adaptation and not an actual score of the quality of semantic matching. 
Using \texttt{BERT} as the initial pre-trained model our results are shown in Table~\ref{tab:table1}.

\begin{table}[h!]
\centering
      \caption{Result for the Domain adaptation with \texttt{BERT} and steps defined in Section~\ref{sec:DA}}
    \label{tab:table1}
    \begin{tabular}{l|c|c} 
      \textbf{Model} & \textbf{Score 1} & \textbf{Score 2}\\
      \hline
      Regular \texttt{BERT} & 0.09 & 0.32 \\
      \texttt{BERT} + Step 1 & 0.10 & 0.33\\
      \texttt{BERT} + Step 1 \& 2 & 0.32 & 0.33\\
    \end{tabular}
\end{table}

The key improvement of the Domain Adaptation comes from fine-tuning Step 2 although we know that regular \texttt{BERT} is not really suited to perform semantic matching.
In fact the low \textbf{Score~1} for regular \texttt{BERT} comes from the fact that both positive and negative matches are high.
Only fine-tuning can help \texttt{BERT} distinguish the negative pairing. 
That is why we also tested the \texttt{all-MiniLM-L6-v2} model using it as is, performed Step~1 and Step~2, and quote the improvement of performing an adaptive pre-training method defined as $\frac{\text{score fine-tune} - \text{score baseline}}{\text{score baseline}}$. 

\begin{table}[h!]
\centering
      \caption{Result for the Domain adaptation with \texttt{all-MiniLM-L6-v2} and steps defined in Section~\ref{sec:DA}}
    \label{tab:table2}
    \begin{tabular}{l|c|c} 
      \textbf{Model} & \textbf{Score 1} & \textbf{Score 2}\\
      \hline
      \texttt{all-MiniLM-L6-v2} & 0.21 & 0.46 \\
      \texttt{all-MiniLM-L6-v2} + Step 1 \& 2 & 0.27 & 0.56\\
      Improvement & 29$\%$ & 22$\%$\\
    \end{tabular}
\end{table}

This shows again the improvement over the baseline.  In addition we tested the impact of the quality of the training/validation dataset by decreasing the score threshold from~$0.7$ down to~$0.6$ when generating the training/validation dataset in Section~\ref{secval} and found that it decreases both scores by $10\%$.
Finally we find that the fraction of masked words in the Mask Language Modelling does not impact significantly the results. By default the fraction is set to 0.15 and we tried 0.2 in the \texttt{BERT} Step 1 and did not find a change in the scoring. 
Overall, using limited data for Step~1 and Step~2 we find that the key effect comes from the fine-tuning as was previously stated in~\cite{SBERT}.

\subsection{Fully unsupervised training with the GPL method}\label{secGPL}
This time we perform fine tuning solely by running the GPL method on the FCA Rulebook corpus and train the GPL method on 10,000 and 20,000 triplets. 
To do so we follow the steps described in \url{github.com/UKPLab/gpl} using the default pre-trained model MSMARCO.  We output samples at each step to check the pseudo-labelling.  
On Step~1, to generate queries from a passage of the corpus to get positive pairs, such as
\begin{example}\ \\
\textup{Selected passage:}
\textit{`The investor providing the capital may choose not to be involved in the running of the venture'}.\\
\\
\textup{Generated queries: }
\begin{itemize}
\item \textit{how are investors involved in a venture}
\item \textit{who is involved in the running of the venture}
\item \textit{what are the investors of a venture}
\end{itemize}
\end{example}

\begin{example}\ \\
\textup{Selected passage:}
\textit{`If a circular submitted for approval is amended a copy of amended drafts must be resubmitted marked to show changes made to conform with FCA comments and to indicate other changes'}.\\
\\
\textup{Generated queries: }
\begin{itemize}
\item \textit{how to amend FCA circular}
\item \textit{what must FCA amendments show}
\item \textit{how to revise circular for approval}
\end{itemize}
\end{example}

\begin{figure}
\begin{center}
\includegraphics[scale=0.4]{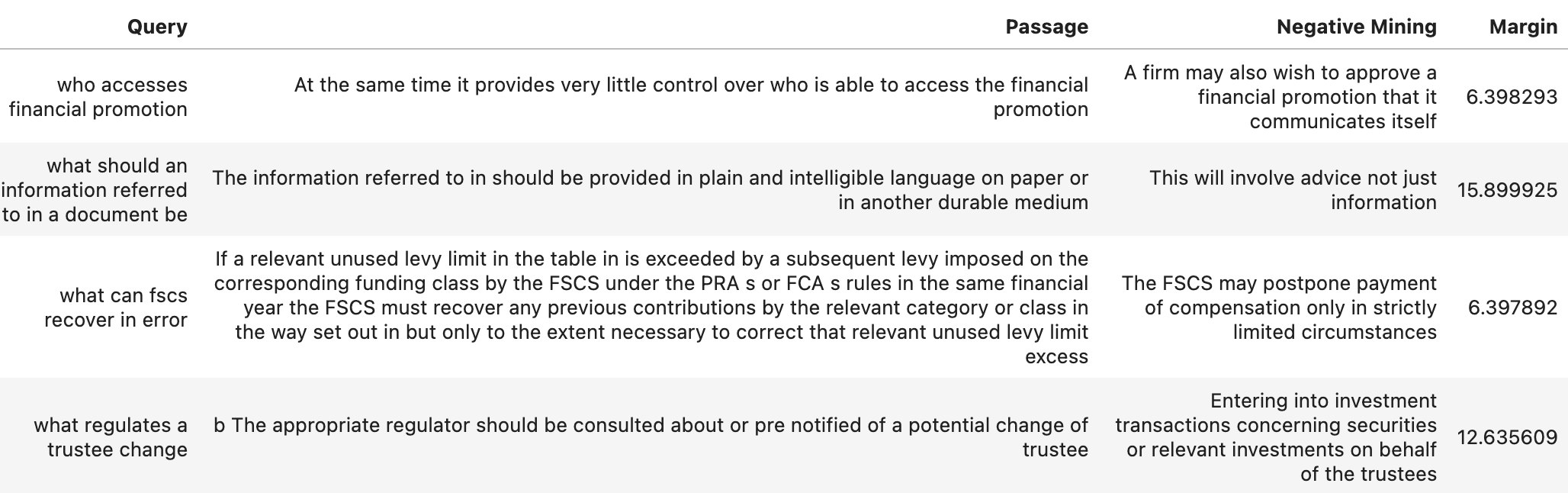}
\caption{Sample of the training dataset generated by the GPL approach from the FCA Rulebook. The Margin is defined in~\eqref{eq:EqMargin}. This data is used to fine-tune the msmarco pre-trained model.}\label{Fig5}
\end{center}
\end{figure}

Clearly some of these positive pairs can be noisy. 
A sample of our generated training datasets containing our triplets of query, positive, negative passages is displayed in Figure~\ref{Fig5}.

\subsubsection{Results}
Similarly to the scoring analysis in Section~\ref{secFT}, 
we see that the GPL fine-tuning does not impact the scores as much as the adaptative pre-training method.  
The scorings themselves are again to be considered as a benchmark from the no-fine-tuning scoring which translate into an improvement of the ratio
$\frac{\text{score fine-tune} - \text{score baseline}}{\text{score baseline}}$.
As expected, GPL fine-tuning is not as efficient as mapping rules to policies, most likely because it did not perform the training over a set of policies but solely on rules.  We also find that performing the training on 10,000 or 20,000 triplets does not impact the results by more than a percent. 

\begin{table}[h!]
  \centering
    \caption{Result for the GPL fine-tuning on the pre-trained MSMARCO model.}
    \label{tab:table3}
    \begin{tabular}{l|c|c} 
      \textbf{Model} & \textbf{Score 1} & \textbf{Score 2}\\
      \hline
      \texttt{msmarco} & 0.14 & 0.79 \\
      \texttt{msmarco} + GPL & 0.15 & 0.84\\
      Improvement & 7$ \%$ &  6$ \%$ \\
    \end{tabular}
\end{table}

\section{Discussion and comparison between GPL and Adaptative pre-training}\label{sec3}

\begin{figure}
\begin{center}
\includegraphics[scale=0.4]{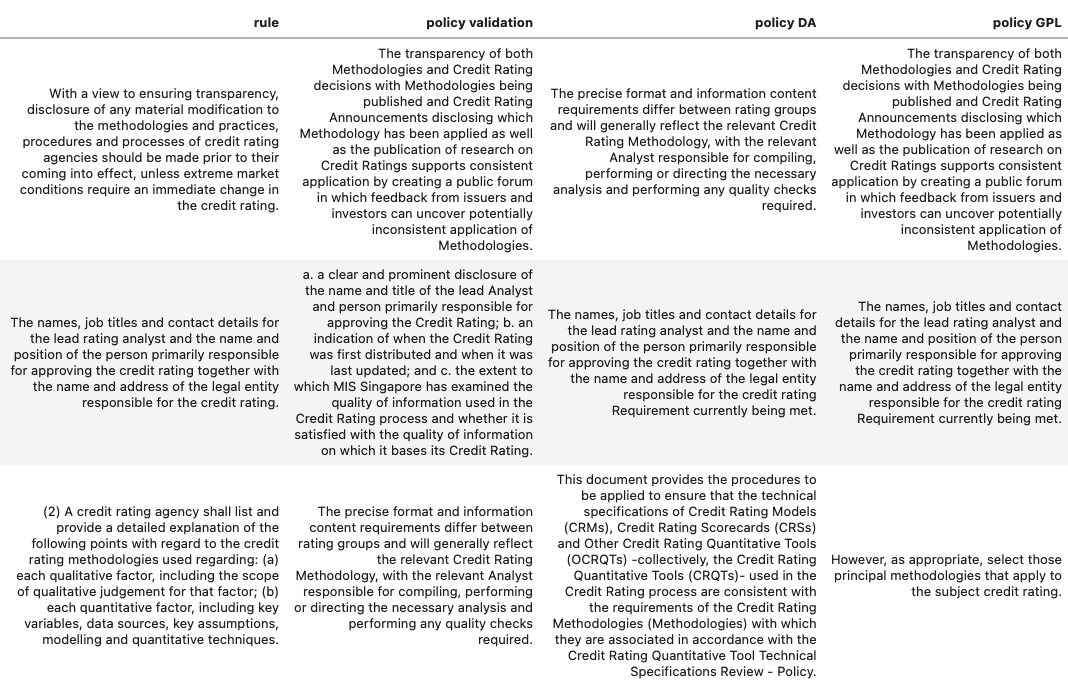}
\caption{Sample comparison of pairing based on cosine similarity from our pseudo validation dataset where we add the best match of the DA and GPL approaches.}\label{Fig7}
\end{center}
\end{figure}

When looking at the validation dataset for the rule and policy match as well as the policy identifies by the Adaptative pre-training (DA) and the GPL fine-tuning, there is no clear way to identify which pairing is the more accurate. 
In fact they all return meaningful pairings.
In Figure~\ref{Fig7} we show a sample where either the DA or the GPL policy match does not agree with the validation policy.  
It seems 
that the validation policy is not always the best match to address the rule. 
In addition, not all the rules have been addressed in our policy sample.  Therefore it is not possible to conclude which approach performs the best matching to our rules. 
The three approaches we investigate in this work are:
\begin{itemize}
\item Mean ensemble of N pre-trained models selecting policies if they are returned by at least $\sqrt{N}$ models with cosine similarity $>0.7$. This is our validation policy defined in Section~\ref{secval}.
\item Model from an Adaptative pre-training (DA), defined in Section~\ref{sec:DA}
\item Model from GPL fully unsupervised training from rules corpus in Section~\ref{secGPL}.
\end{itemize}

It is likely that a combination of these three approaches should provide the best pairing.  Overall it is quite remarkable that even in the absence of a training dataset, we manage to perform an Adaptative pre-training that improves the scores. This can be helpful as an alternative or a complement to the GPL approach. 

\section*{Acknowledgements}
The authors would like to thank Rohini Gupta and Amit Madhar from Finreg-E (\url{https://finreg-e.com}), 
without whom this project would not have been possible.
IA and AJ are supported by the
Innovate UK Smart Grant \textit{`Finreg-E / Natural Language Processing for Financial Regulation'}. DG is supported by the ESRC Grand Union Doctoral Training Partnership Grant and the Oxford Man Institute of Quantitative Finance.

\bibliography{sn-bibliography}

\end{document}